%% file: main_camera_ready.tex
\documentclass[10pt,twocolumn,letterpaper]{article}

\usepackage[pagenumbers]{iccv} 


\definecolor{iccvblue}{rgb}{0.21,0.49,0.74}
\usepackage[pagebackref,breaklinks,colorlinks,allcolors=iccvblue]{hyperref}
\usepackage{graphicx} 
\usepackage{booktabs} 
\usepackage{float}
\definecolor{darkpurple}{RGB}{128, 0, 128} 
\definecolor{lightlydarkyellow}{RGB}{204, 204, 0}
\definecolor{darkgreen}{RGB}{0, 100, 0}
\definecolor{darkred}{RGB}{139, 0, 0}
\definecolor{darkpink}{RGB}{231, 84, 128}
\definecolor{highlightred}{RGB}{255, 102, 102}
\definecolor{mypink}{RGB}{180, 100, 190} 
\usepackage[utf8]{inputenc}
\usepackage{tcolorbox}
\usepackage{multirow}
\tcbuselibrary{listings, breakable}

\usepackage{mathtools}

\usepackage{algorithm}
\usepackage{algpseudocode}
\usepackage{multirow}
\usepackage{pifont}

\algrenewcommand{\Return}{\State\algorithmicreturn~}

\newtcolorbox[list inside=prompt,auto counter,number within=section]{prompt}[1][]{
    colbacktitle=black!60,
    coltitle=white,
    fontupper=\footnotesize,
    boxsep=5pt,
    left=0pt,
    right=0pt,
    top=0pt,
    bottom=0pt,
    boxrule=1pt,
    title={Prompt \thetcbcounter: #1},
    breakable,
}
\def\paperID{3327} 
\def\confName{ICCV}
\def\confYear{2025}

\newcommand{\ours}[0]{{DWIM}}

\title{DWIM: Towards Tool-aware Visual Reasoning via Discrepancy-aware Workflow Generation \& Instruct-Masking Tuning}

\author{
Fucai Ke$^{1,2}$ \quad
Vijay Kumar B G$^{3}$ \quad
Xingjian Leng$^{4}$ \quad
Zhixi Cai$^{2}$ \quad
Zaid Khan$^{5}$ \\
Weiqing Wang$^{2}$ \quad
Pari Delir Haghighi$^{2}$ \quad
Hamid Rezatofighi$^{2}$ \quad
Manmohan Chandraker$^{3,6}$ \\
{\tt\small
$^1$Building 4.0 CRC }
{\tt\small
$^2$Monash University }
{\tt\small
$^3$NEC Labs America }
{\tt\small
$^4$ANU }
{\tt\small
$^5$UNC Chapel Hill }
{\tt\small
$^6$UC San Diego 
}\\
{\tt\small \url{https://pokerme7777.github.io/DWIM.github.io/}}
}

\begin{document}
\twocolumn[{
\begin{center}
\maketitle

\captionsetup{type=figure}
\vspace{-1em}
\includegraphics[width=0.8\textwidth]{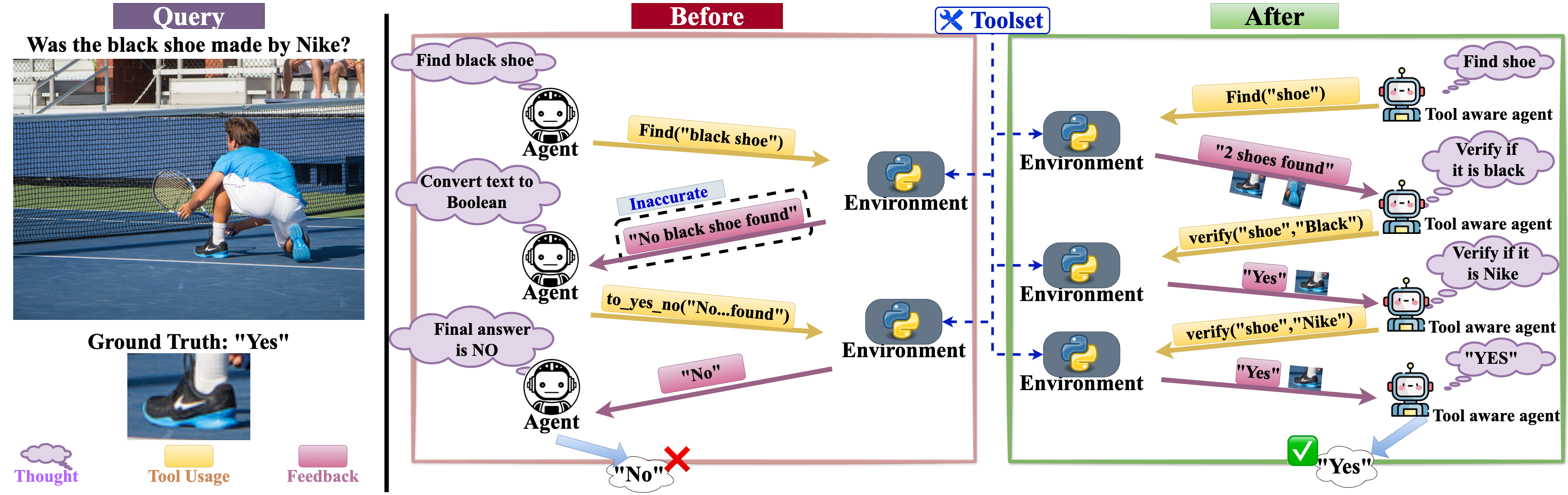}
\vspace{-0.5em}
\captionof{figure}{Comparison of the existing agent (Before) and our tool-aware agent (After). Both follow logically valid workflows with the same toolset, but our method improves tool selection and usage, minimizing tool-induced errors and ensuring more accurate, efficient execution.}
\label{fig:teaser}
\end{center}
}
]

\input{sec/0_abstract}    
\input{sec/1_new_intro}
\input{sec/2_related_work}
\input{sec/3_method}

\input{sec/4_experiment}

\input{sec/5_conclusion}
\section*{Acknowledgments}
This work was supported by Building 4.0 CRC and the Commonwealth of Australia through the Cooperative Research Centres Program. It was also partially funded by the DARPA ANSR program (FA8750-23-2-1016) and the ARC DECRA program (DE250100032).

{
    \small
    \bibliographystyle{ieeenat_fullname}
    \bibliography{main}
}

\input{sec/X_suppl}

\end{document}

%% file: sec/0_abstract.tex
\begin{abstract}\label{sec:abstract}Visual reasoning (VR), which is crucial in many fields for enabling human-like visual understanding, remains highly challenging. Recently, compositional visual reasoning approaches, which leverage the reasoning abilities of large language models (LLMs) with integrated tools to solve problems, have shown promise as more effective strategies than end-to-end VR methods. However, these approaches face limitations, as frozen LLMs lack tool awareness in VR, leading to performance bottlenecks. While leveraging LLMs for reasoning is widely used in other domains, they are not directly applicable to VR due to limited training data, imperfect tools that introduce errors and reduce data collection efficiency in VR, and challenging in fine-tuning on noisy workflows. To address these challenges, we propose DWIM: i) Discrepancy-aware training Workflow generation, which assesses tool usage and extracts more viable workflows for training; and ii) Instruct-Masking fine-tuning, which guides the model to only clone effective actions, enabling the generation of more practical solutions. Our experiments demonstrate that DWIM achieves state-of-the-art performance across various VR tasks, exhibiting strong generalization on multiple widely-used datasets. 
\vspace{-1em}
\end{abstract}

%% file: sec/1_new_intro.tex
\section{Introduction}
\label{sec:newintro}

This paper addresses the problem of visual reasoning (VR), which involves constructing detailed visual scene representation and  reasoning about it in steps, similar to human cognition. VR involves interpreting and analyzing visual information in response to textual queries or prompts ~\cite{amizadeh2020neuro, ke2024hydra, cai2024neusis, NEURIPS2023_b14cf0a0} and encompasses a diverse range of tasks, including, but not limited to, visual commonsense reasoning~\cite{hudson2019gqa}, external-knowledge visual question answering~\cite{marino2019okvqa,schwenk2022aokvqa}, vision-language compositionality understanding~\cite{hsieh2024sugarcrepe, thrush2022winoground}, visual adversarial sample answering~\cite{li2024naturalbench} visual grounding~\cite{kazemzadeh2014refcoco} and complex counting~\cite{acharya2019tallyqa}. Beyond basic perception, VR enhances data interpretation by integrating reasoning with visual understanding, making it crucial in fields such as multimodal learning, autonomous driving, unmanned systems, and medical image analysis~\cite{cai2024neusis, cui2024survey, cui2024drive, zhan2020medical}. However, despite its significant progress, VR remains highly challenging due to the inherent complexity of reasoning and the vast diversity of tasks~\cite{liu2024improvedllava15, li2023blip2, zhu2023minigpt, hurst2024gpt4o}.

The most common VR solutions rely on end-to-end vision-language models~\cite{liu2024visualllava, liu2024improvedllava15}, which require large annotated datasets and significant computational resources, limiting their scalability and efficiency. Recently, compositional approaches, also known as workflow-based~\cite{zeng2023flowmind, wu2024stateflow} or tool-aware methods~\cite{wang2024voyager, qin2024tool}, have emerged, leveraging frozen LLMs as planners to break down complex tasks into smaller, more manageable sub-tasks. By generating structured tool-utilization workflows and precise actions, these methods improve performance~\cite{gupta2023visprog, suris2023vipergpt, lu2023chameleon, NEURIPS2023_b14cf0a0, ke2024hydra, shen2024hugginggpt, cai2024neusis, ma2024gerea}. However, their effectiveness remains limited, as most rely solely on the pre-trained knowledge of frozen LLMs.

Notably, the reasoning capabilities of LLMs have also been leveraged in other domains, such as mathematical reasoning, text-based games, and question answering, through similar step-by-step reasoning methods~\cite{yin2024agentlumos, zhu2024knowagent, qiao2024autoact, hu2022lora, liu2024visualllava, zhu2023minigpt, khan2024visrep, yao2022react, pmlrcodeact,gao2025multimodalagenttuningbuilding}. 
In these domains, LLM agents are fine-tuned as domain experts rather than used in their frozen state. This is because frozen models are not explicitly trained to understand tool capabilities or generate effective tool-utilization workflows. To address this, approaches typically rely on large datasets containing annotated workflows, or leverage the inherent reasoning abilities of LLMs to generate workflows, selecting those that produce correct outcomes for training~\cite{shinn2024reflexion, wu2025avatar, yang2024embodied, yin2024agentlumos}. LLMs are then trained on these workflows using supervised fine-tuning (SFT) or reinforcement learning (RL) to enhance their reasoning capabilities (\eg ~\cite{yin2024agentlumos, zhu2024knowagent, zhai2025fine}).

\begin{figure}[t]
\vspace{-2pt}
\begin{center}
  \includegraphics[width=0.8\linewidth]{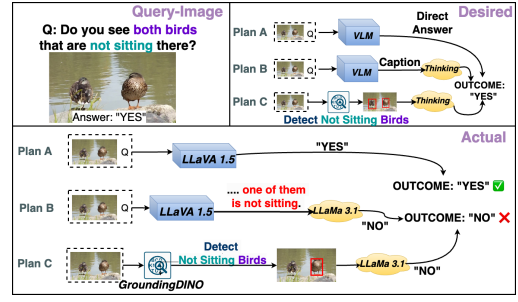} 
\end{center}
\vspace{-1em}
\caption{Tools are not always reliable and may occasionally provide incorrect information. Consequently, workflows expected to yield correct answers may fail due to tool-related inaccuracies.}
\label{fig:vfm_problem}
\vspace{-1em}
\end{figure}

Similarly, while training LLMs is crucial for VR~\cite{khan2024visrep}, applying the ideas from the aforementioned domains naively are not suitable due to the broader range of VR tasks and the relatively smaller dataset sizes. Consequently, most compositional VR approaches (\eg ~\cite{you2023idealgpt, gupta2023visprog, suris2023vipergpt, ke2024hydra}) avoid training LLMs and instead rely on frozen language models. However, frozen LLMs limit overall performance~\cite{khan2024visrep}.
Additionally, in the previously mentioned domains, action execution is generally reliable, as tools (\eg, calculators, in-game functions, and search engines) operate based on deterministic algorithms. However, compositional visual reasoning heavily relies on deep learning models, which are more prone to inaccuracies and factual errors during execution.
As illustrated in Figure~\ref{fig:vfm_problem}, many logically sound workflows should yield correct answers (Desired block), but environmental feedback (\ie, tool errors) cause some to fail in practice (Actual block). This limits the generation of valid workflows, worsening the data shortage for training, as faced by \cite{khan2024visrep}. Additionally, filtering workflows solely based on the final outcome and cloning entire workflows during training, without identifying tool usage effectiveness, may cause the model to learn noise (\ie, ineffective tool usage) in naive SFT.

In this paper, we introduce DWIM, which consists of: (i) Discrepancy-aware training Workflow generation to assess tool usage, discover more effective workflows, and optimize data utilization; and (ii) Instruct-Masking, a fine-tuning method that enhances tool awareness, improving task planning, and learns effective tool usage from noisy workflows.
Our discrepancy-aware training workflow generation, integrated with an agentic framework, enables LLMs to recognize ineffective tool usage by detecting factual errors or unexpected information in environmental feedback. By conditioning on the ``answer,'' the LLM recognizes discrepancies (\eg, factual errors) between execution feedback and expected outputs, describes them, and refines tool usage in following steps. This refinement enables the agent to explore alternative tool usages, which increase workflow success rates and ultimately yield more workflows for training.

Once workflow generation is complete, we fine-tune the LLM using instruct-masking. Treating a workflow as a sequence of actions (\eg, task planning in text and tool usage in Python), we iteratively mask effective actions (\eg, tool usage) at the semantic level based on prior effectiveness assessments and instruct the model to predict them. 
Ineffective tool usage and its corresponding next step (\ie, discrepancy description and refinement thought) in the unmasked parts help the model recognize failed tool usage cases, while masking facilitates workflow augmentation. Meanwhile, instruction ensures the LLM clones only effective actions rather than blindly clone every step, as in naive SFT.

We evaluate our proposed methods on several popular VR datasets and compare with recent models, showing state-of-the-art performance.
In summary, the key contributions of this work are as follows:
\begin{enumerate}
    \item We introduce \textit{discrepancy-aware training workflow generation}, the first known method to assess VR tool usage effectiveness, refine actions with alternatives, and increase workflow success rates, thereby enhancing training data utilization.

    \item We propose an \textit{instruct-masking fine-tuning}, which helps the model understand tool failures by exposing failure cases in the unmasked parts while instructing it to learn effective tool usage in masked part, expanding the trainable workflow size and enhancing tool awareness.

    \item We conduct extensive experiments to demonstrate the advantages of our proposed method, highlighting its potential to operate effectively with reduced dependency on human prompt engineering.

\end{enumerate}

\begin{table*}
\centering
\caption{Difference between our work and similar compositional VRs.}
\vspace{-1em}
\label{tab:diff_CVR}
\scalebox{0.55}{\begin{tabular}{cccccccc}
\midrule
\rowcolor[gray]{0.95} & Supervision & Multi-turn Framework & Workflow Refinement & Improves LLM & Assess Tool Usage in Training & Improve Data Utilization\\
\midrule
DWIM(ours) & \textcolor{darkgreen}{YES} & \textcolor{darkgreen}{YES} & \textcolor{darkgreen}{YES} & \textcolor{darkgreen}{YES} & \textcolor{darkgreen}{YES} & \textcolor{darkgreen}{YES} \\

HYDRA~\cite{ke2024hydra}& \textcolor{darkgreen}{YES} & \textcolor{darkgreen}{YES} & \textcolor{darkgreen}{YES} & \textcolor{darkred}{NO} & \textcolor{darkred}{NO} & \textcolor{darkred}{NO} \\

VisRep~\cite{khan2024visrep}& \textcolor{darkgreen}{YES} & \textcolor{darkred}{NO} &  \textcolor{darkred}{NO} & \textcolor{darkgreen}{YES} & \textcolor{darkred}{NO} & \textcolor{darkred}{NO} \\

ViperGPT~\cite{suris2023vipergpt}& \textcolor{darkred}{NO} & \textcolor{darkred}{NO} & \textcolor{darkred}{NO} & \textcolor{darkred}{NO} & - & - \\

VisProg~\cite{gupta2023visprog} & \textcolor{darkred}{NO} & \textcolor{darkred}{NO} & \textcolor{darkred}{NO} & \textcolor{darkred}{NO} & - & - \\

IdealGPT~\cite{gupta2023visprog} & \textcolor{darkred}{NO} & \textcolor{darkred}{NO} & \textcolor{darkred}{NO} & \textcolor{darkred}{NO} & - & - \\

Chameleon~\cite{lu2023chameleon} & \textcolor{darkred}{NO} & \textcolor{darkred}{NO} & \textcolor{darkred}{NO} & \textcolor{darkred}{NO} & - & - \\

\bottomrule
\end{tabular}}
\vspace{-1.5em}
\end{table*}

%% file: sec/2_related_work.tex
\section{Related Work}
\label{sec:related_work}

\subsection{Compositional Visual Reasoning Approach}
The compositional visual reasoning approach leverages powerful LLMs to solve visual reasoning tasks \cite{suris2023vipergpt,you2023idealgpt,khan2024visrep,ke2024hydra,gupta2023visprog,cai2025naver, ma2025drvideo}.
In this strategy, LLMs function as planners, reasoners or program generators to break down complex tasks into manageable subtasks and generate corresponding action plans through their chain-of-thought reasoning abilities~\cite{jiang2024diem,hu2024visualdistillation,mitra2024compositionalcot}. The action plan may consist of high-level text descriptions outlining the task goal or fine-grained actions in text, symbols, or even Python code for logical operations~\cite{brown2020languagegpt3, chowdhery2023palm, kojima2022large, huang2022language}. Based on the generated action plan, execution is carried out by leveraging various perception models or external tools, treating other end-to-end (E2E) models as tools (\eg, ~\cite{bai2023qwen,li2022grounded, liu2023groundingdino, liu2024visualllava}) and integrating them into the tool library. For example, VisProg~\cite{gupta2023visprog} and ViperGPT~\cite{suris2023vipergpt} leverage GPT models~\cite{brown2020languagegpt3} to generate code for perception, question answering, and task-solving. Chameleon~\cite{lu2023chameleon} uses GPT-3.5~\cite{brown2020languagegpt3} to generate a tool utilization sequence based on a provided template for task execution, while IdealGPT~\cite{you2023idealgpt} employs a vision-language model to answer each sub-question generated by GPT-3.5, subsequently aggregating the results through reasoning. 

However, these compositional methods face two key limitations: (1) they rely on frozen LLMs, which lack the ability to decompose problems into solvable subproblems that effectively leverage the provided tools. In-context learning examples alone are insufficient for enabling LLMs to fully grasp these tools' capabilities to complete tasks~\cite{qiao2024autoact, xiong2024watch}; and (2) they employ a single-turn framework instead of a multi-turn agentic approach, which confines LLMs to initial planning without iterative feedback from the environment, thereby limiting overall performance~\cite{ke2024hydra}.

Recently, VisRep~\cite{khan2024visrep} introduced a method that leverages existing annotations to generate a coarse reward signal for visual reasoning tasks. It uses naive SFT to improve the LLM's visual program synthesis ability. However, VisRep's performance is limited by its single-turn framework, training workflow generation, and fine-tuning approach, which collects only a small portion of training workflows and clones full workflows, including ineffective actions.
In contrast, another recent SOTA method, HYDRA~\cite{ke2024hydra}, introduces a multi-turn incremental reasoning agentic framework. Its planner and reasoner modules use an LLM to generate instructions and executable code, while a RL agent makes high-level decisions based on feedback, allowing HYDRA to adjust actions for greater accuracy and effectiveness. However, as the planner itself is untrained, it may still produce incorrect next-step suggestions and lacks the ability to refine them. The differences between compositional VRs are illustrated in Table~\ref{tab:diff_CVR}.

\subsection{Workflow-based and Tool-aware LLM Agent}\label{subsec:agent_training}
LLM agents offer advantages in affordability, transparency, and reproducibility for complex tasks~\cite{xiong2024watch, yin2024agentlumos, qiao2024autoact, zhao2024expel, yao2022react}. 
Incremental reasoning agent framework such as~\cite{yao2022react, yuancraft, wang2024trove, koo2024proptest, yin2024agentlumos, qiao2024autoact, zhu2024knowagent, xiong2024watch, qiao2024agent}, apply workflow-based and tool-aware LLM agents across language understanding, math, and beyond. To preparing for workflow training, these methods reply existing datasets or generate logically sound workflows using GPT-4~\cite{achiam2023gpt4} or other LLMs, discarding outputs that yield incorrect results. This process is referred to in this paper as the “\textbf{standard}” training workflow generation method. However, it is unsuitable for visual reasoning tasks, as it fails to account for common errors arising from imprecise tools during interactions. Additionally, while their supervised fine-tuning approach is effective in certain domains, it may unintentionally reinforce incorrect tool usage in visual reasoning tasks.

\subsection{Masking and Instruction Tuning}\label{subsec:masking}
In BERT~\cite{kenton2019bert}, masking refers to randomly replacing words or tokens in a sentence with a mask token (\eg, \texttt{[MASK]}) and training the model to predict these hidden tokens based on surrounding context.
Instruction tuning, on the other hand, is a fine-tuning process that trains LLMs to follow specific prompts using diverse instruction-output pairs. This method, which includes machine-generated instruction-following data, has been shown to enhance zero-shot performance across a range of tasks, especially in natural language processing and visual language models~\cite{brown2020languagegpt3,ouyang2022training,liu2024improvedllava15,daiInstructBLIP,kenton2019bert}. Our method, termed instruct-masking fine-tuning, is inspired by these approaches.

%% file: sec/3_method.tex
\section{Methodology}
\label{sec:methodology}
To develop a more tool-aware LLM for visual reasoning, we propose DWIM: Discrepancy-aware training Workflow generation to assess tool usage, discover effective workflows, and optimize data utilization, along with Instruct-Masking fine-tuning to enhance tool awareness using collected noisy workflows.

\subsection{Task Formulation}
This subsection outlines the VR task formulation. In a visual reasoning task, let $v$ be a visual input and  $q$ be a textual query about $v$, thus for each task $\xi = \{(v, q), y\}$, there exists a visual-textual query pair $(v,q)$ corresponding to the answer $y$. When using the compositional visual reasoning method, there is a workflow $\boldsymbol{\omega} = \{\omega_{1:T}\}$ that contains each generated  action $\omega$ from an agentic LLM $\pi_\theta$. For any step $t$ satisfying $1 \leq t \leq T$, we have:
\vspace{-8pt}
\begin{align}\label{eq:generate_workflow}
    \pi_\theta(\omega_t|e_0) &= \prod_{t'=1}^{t}\pi_\theta(\omega_{t'}|e_{t'-1}), \\
    e_{t} &= \{e_{t-1}, \phi(\omega_{t})\},
\end{align}
where $e_t$ is a combination of previous environmental feedback and the new feedback received after the interaction with action $\omega_t$. $\phi$ represents the function that maps each action taken by the agent to the corresponding feedback from the current environment context. The environmental feedback $e_t$ includes action execution outputs and error messages based on the query pair $(v,q)$. The environment information for each task is given by $e_0 = \{(v, q), \delta\}$, where $\delta$ denotes the tool library documentation.
Thus for each task, the answer prediction is approximately
\vspace{-8pt}
\begin{align}\label{eq:prediction}
    \hat{y} &:= \phi(\omega_{T})\\
    \omega_{T} & \sim  \pi_{\theta}(\omega_{T} | e_{T-1})\\
    \pi_{\theta}(\omega_{T} | e_{T-1}) &= \pi_{\theta}(\omega_{T} | \{e_0,\phi(\omega_1),\dots,\phi(\omega_{T-1})\})\label{qe:standardmethod}
\end{align}

For training, our objective is to optimize the parameters $\theta$ of the agentic LLM $\pi$ to accurately generate the workflow. Building on these, we propose our workflow generation and fine-tuning methods.


\subsection{Workflow Generation with Agentic Framework}
\label{subsection:DWIM}
This subsection introduces our proposed training workflow generation method for workflow collection, including discrepancy-aware recognition of ineffective tool usage, as well as the dynamic agentic framework designed for VR.

Our discrepancy-aware training workflow generation relies on the Auto-Exploring Agentic Framework, which is inspired by ReAct~\cite{yao2022react} and CodeAct~\cite{pmlrcodeact}. However, unlike these methods, our proposed solution features a more interactive and dynamic agentic structure to enable discrepancy-aware recognition at each step, considering potential errors in environmental feedback during training workflow generation.
This framework necessitates multi-turn interactions, moving beyond the traditional one-step planning framework (\eg, \cite{suris2023vipergpt, khan2024visrep, you2023idealgpt}), as illustrated in Figure~\ref{fig:framework}.

In a single-turn framework, $\hat{y}$ is approximated as:
\vspace{-8pt}
\begin{align}
    \hat{y} &:= \phi(\omega_1) \\
    \omega_1 &\sim \pi_\theta(\omega_1 | e_0),
\end{align}
where $e_0$ is the initial environment feedback. Thus, the prediction $\hat{y}$ is limited by $\pi_\theta$ and $e_0$ alone due to the single-turn setup.
In contrast to existing single-turn agentic framework, multi-turn agentic framework~\cite{yao2022react, pmlrcodeact} offer the advantage of incorporating aggregated environment information $\{e_{1},\dots,e_{t-1}\}$ to enable incremental reasoning. These information $e$ are grounded in the environmental context $\{(v,q),\delta\}$, enabling the model to iteratively refine its action generation process. 

As illustrated in Figure~\ref{fig:framework} (Supplementary), there are three types of generated outputs inspired by CodeAct~\cite{pmlrcodeact}, $\omega_t \in \{\text{\textlangle Thought\textrangle}, \text{\textlangle Code\textrangle}, \text{\textlangle Done\textrangle}\}$, corresponding to thinking, acting and sensing, and mission completion, respectively. The \textlangle Thought\textrangle-type action $\omega_t$ enhances the reasoning process by analyzing the provided environmental information $e_t$ to determine the optimal next step, either proceeding or refining the action, while articulating the rationale in natural language. When a \textlangle Code\textrangle-type action (\ie, tool usage) $\omega_t$ is generated, the agentic LLM explores  $\phi(*)$, utilizing (perception) tools to gather additional necessary information. This new environmental feedback is appended to the existing environment information, updating $e$ incrementally as $e_t=\{e_{t-1}, \phi(\omega_t)\}$ to support incremental reasoning. The exploration process is achieved by generating Python code that is executed using predefined Python APIs. Once $e_{T-1}$ contains sufficient information and the agentic LLM has answered the query $q$, a \textlangle Done\textrangle-type action $\omega_T$ is generated to conclude the task.

\begin{figure}[t]
\begin{algorithm}[H]

\caption{\small Discrepancy-aware training workflow generation }
\begin{algorithmic}[1]
\setlength{\parskip}{0pt}
\setlength{\floatsep}{0pt}
\setlength{\textfloatsep}{0pt}
\setlength{\intextsep}{0pt}
\footnotesize
\Require{$ v, q, \phi, \pi_\theta, \textcolor{highlightred}{y}, D_{\boldsymbol{\omega}}$}
\State $e \gets (v, q); t \gets 1; \boldsymbol{\omega} \gets \{\}$ \Comment{Initialization}
\While{not final}
    \State $\omega \gets \{\pi_\theta, \boldsymbol{\omega}, e, \textcolor{highlightred}{y}\}; \boldsymbol{\omega}.\text{append}(\omega)$ \Comment{Generate and store action}
    
    \State $e \gets \{\phi, \omega, e\}$\Comment{Update environment Info}

    \If{discrepancy in $(e, y)$} \Comment{\textcolor{highlightred}{Discrepancy-aware Recognition}}
        \State $\textcolor{highlightred}{\omega_{\text{Rethink}}} \gets \{\pi_\theta, \boldsymbol{\omega}, e, y\}$\Comment{\textcolor{highlightred}{Generate ``Rethink''}}
        \State $\boldsymbol{\omega}.\text{append}(\textcolor{highlightred}{\omega_{\text{Rethink}}} ); t \gets t + 1$\Comment{\textcolor{highlightred}{Store ``Rethink''}}
    \EndIf
    \State $t \gets t + 1$
\EndWhile
\If{$y= \hat{y} \gets \phi(\omega_{t})$} 
    \State $D_{\boldsymbol{\omega}}.append(\boldsymbol{\omega}\}$\Comment{Collect workflow}
\EndIf\\
\textbf{return} $D_{\boldsymbol{\omega}}$

\label{algorithm1}
\end{algorithmic}
\end{algorithm}
\vspace{-10mm}
\end{figure}

\begin{figure*}[t]
\begin{center}
  \includegraphics[width=0.7\linewidth]{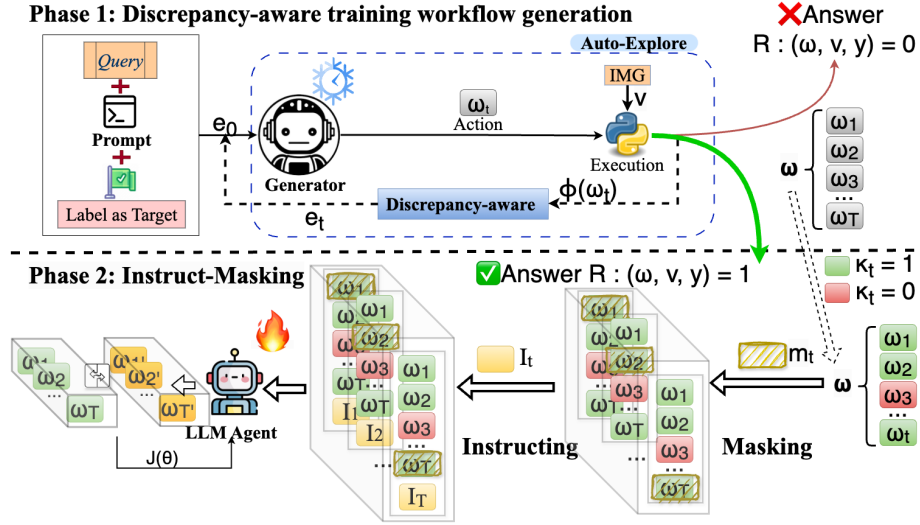} 
\end{center}
\vspace{-1.5em}
\caption{Overview of discrepancy-aware training workflow generation and instruct-masking process} 
\label{fig:method}
\vspace{-1.5em}
\end{figure*}

\textbf{Discrepancy-aware Training Workflow Generation.} 
Leveraging the agentic framework, we design a new training workflow generation method to enhance workflow creation and action assessment, with the algorithm presented in Algorithm 1.
During the training workflow generation phase, the answer $y$ is provided as prior knowledge, which modifies the policy $\pi_\theta$ as follows:
\vspace{-8pt}
\begin{equation}
    \pi_\theta(\omega_t | e_0, y) = \prod_{t'=1}^{t} \pi_\theta(\omega_{t'} | e_{t'-1}, y).
\end{equation}

The \textbf{``standard''} workflow generation method, as discussed in Section~\ref{subsec:agent_training}, generates a workflow using $\pi_\theta(\omega_T | e_{T-1})$, which only accepts $e_{T-1}$ as a pre-condition without $y$. 
A multi-turn framework with $T$-turn interactions produces $\omega_T$ based on ${e_{T-1}}$, whereas a single-turn framework relies solely on ${e_{0}}$. 
Given that $y$ is correct, a richer prior and consistent policy increase the likelihood of executing a correct action. 

Simply put, if each $e_t$ is potentially inaccurate, conditioning on $y$ enables the LLM to recognize discrepancies between the ideal outcome and actual feedback. We prompt the LLM to assess each step’s effectiveness by identifying such discrepancies (\eg, factual errors from tool failures). If discrepancies are found, LLM generates a \textit{``Rethink”} \text{\textlangle Thought\textrangle} action $\omega_{\text{Rethink}}$, which provides a natural language description of the discrepancy and a suggested alternative as the next step.
The red parts in Algorithm 1 highlight the advantage of the designed discrepancy-aware training workflow generation compared to the ``standard'' methods.
Relying on new generation policy, we can then collect a dataset of workflow $D_{\boldsymbol{\omega}}$ with assessment result of each step action for instruct-masking fine-tuning.

\subsection{Instruct-Masking Fine-tuning}
Based on the collected workflows, we fine-tune the model on these noisy workflows using stepwise assessment results. Our goal is to use the collected dataset \( D_{\boldsymbol{\omega}} \) to improve the policy \( \pi_\theta \) by tuning the parameters \( \theta \). Existing methods~\cite{khan2024visrep, yin2024agentlumos, ke2024hydra} directly use a binary-valued reward function \( R: (\boldsymbol{\omega}, v, y) \rightarrow \{0,1\} \), which disregards the effectiveness of individual actions \( \omega_t \). However, our discrepancy-aware training workflow generation method can store intermediate feedback \( e \) and enables action flagging to identify effective actions, as detailed in Section~\ref{lab_action_flagging}.
Each action \( \omega_t \) is flagged as effective (\( \kappa_t = 1 \)) or ineffective (\( \kappa_t = 0 \)), resulting in a flagged action sequence.

Then each effective action is masked and the LLM is instructed to reproduce it iteratively. The loss is computed between the original and the LLM-generated action, allowing the model to focus on learning effective tool usage and planning strategies, rather than memorizing entire noisy workflows.
In contrast to BERT's masking~\cite{kenton2019bert}, we mask the action in a workflow at the semantic level instead of token level, as illustrated in phase 2 of Figure~\ref{fig:method}. The instruct approach is inspired by instruction-tuning~\cite{daiInstructBLIP,liu2024visualllava,brown2020languagegpt3} and incorporates the concept of an end-of-sequence token~\cite{weiss2017sequence,vaswani2017attention,ke2024hitskt,hirschberg2015advances}. 

\textbf{Instruct-Masking.}
We define $m_t^\xi$ as a context-level mask for $\omega_t^\xi$ when \( \kappa_t = 1 \) in task $\xi$. The instruction $I_t^\xi$ corresponds to instructing the model to regenerate the masked action $\omega_t$ instead of proceeding to the next step. We then transition $\boldsymbol{\omega}^\xi$ into $d_t^\xi = \{\omega_{1:t-1}^\xi, m_t^\xi, \omega_{t+1:T}^\xi, I_t^\xi\}$ and the new instruct-masking dataset can de denoted by $D_{M}$. Then, we apply behavioral cloning by minimize the reward-weighted loss 
\vspace{-8pt}
\begin{equation}
    J(\theta) = \mathbb{E}_{(\omega,e)} \sim D_{M}[R(\boldsymbol{\omega}, v)\mathcal{L}_{\text{NLL}}(p,q;\theta)],
\end{equation}
where $\mathcal{L}_{\text{NLL}}(p,q;\theta)$ is the negative log-likelihood loss
\vspace{-8pt}
\begin{equation}
    \mathcal{L}_{\text{NLL}}(p, q; \theta) = -\mathbb{E}_{(\omega, e) \sim D_{M}} \left[ \sum_{t=1}^{T} \log \pi_\theta(\omega_t | d_t, e) \right].
\end{equation}
\vspace{-3em}

%% file: sec/4_experiment.tex
\section{Experiments and Results}
\label{sec:experiment}
We evaluate DWIM on multiple datasets through extensive experiments, comparing it to SOTA. Experiments show that DWIM enhances LLM tool usage in both effectiveness and efficiency while demonstrating strong robustness.

\textbf{Implementation Details.}
All experiments are conducted with $4$ RTX A6000 GPUs. For all experiments requiring open-source LLMs, we use LLaMa-3.1-8B-Instruct as the LLM backbone~\cite{dubey2024llama}. We use vLLM~\cite{kwon2023efficientvllm} for accelerating the agent exploration process. To encourage the agent to explore various tool-usage actions, a temperature of $0.8$ is set for LLM generation. 
For model fine-tuning, we leverage Low-Rank-Adaptation (LoRA)~\cite{hu2022lora} with a rank $r=64$ and a scaling factor $\alpha=16$. The training schedule uses a cosine annealing scheduler with a peak learning rate of $3 \times 10^{-5}$ and a warm-up ratio of $0.05$. The model is trained over $6$ epochs with a global batch size of 128.

During evaluation, we apply greedy sampling for generative models to ensure the reproducibility of results. We strictly adhere to the evaluation protocols established by previous works~\cite{khan2024visrep,ke2024hydra, gupta2023visprog, suris2023vipergpt}, including dataset splits and the official baseline codebase. The same tool library and language backbone were used, except for HYDRA, which employs GPT-4o due to the unavailability of GPT-3.5. Furthermore, to ensure optimal performance reproduction, we adjust the total number of in-context learning examples to $10$ based on the provided examples in the official code. The choice of ten examples is supported by our study, which will be presented in Section 4.4 and Figure~\ref{fig:shot_diff}. As a result, the reported performance may vary slightly, being either higher or lower than the originally reported results in the HYDRA and VisRep papers.

\textbf{Task.}
The visual reasoning task primarily encompasses visual commonsense reasoning (VCR), external-knowledge visual question answering (EKVQA), vision-language compositionality understanding (VLCU), visual adversarial samples answering (VASA), grounding detection (GD) and complex counting questions (CCQ). 

\textbf{Dataset and Metric.} In the method comparison experiment, we train and test DWIM and baseline models separately across six datasets~\cite{hudson2019gqa, marino2019okvqa, hsieh2024sugarcrepe, li2024naturalbench, kazemzadeh2014refcoco, acharya2019tallyqa}, each representing a distinct task, to evaluate their performance. We evaluate generalization on four datasets to ensure fairness, as some require model-specific adaptations that could introduce bias. Each method follows a consistent training and evaluation framework using the same inference setup across training set~\cite{hudson2019gqa} and all four evaluation sets~\cite{marino2019okvqa, schwenk2022aokvqa, thrush2022winoground, hsieh2024sugarcrepe}.
Following previous work~\cite{khan2024visrep, ke2024hydra, suris2023vipergpt}, we use the same dataset split settings on TallyQA~\cite{acharya2019tallyqa}, which was not covered in prior studies, creating a train set of $1,000$ samples and test sets of $500$ samples. Performance is assessed using accuracy (ACC)~\cite{suris2023vipergpt, you2023idealgpt, lu2023chameleon, ke2024hydra, khan2024visrep}. In the case of the VASA dataset (\ie NaturalBench~\cite{li2024naturalbench}), we report group-level ACC rather than single-question ACC, in alignment with the guidelines provided by the dataset authors. For the grounding detection task, we use intersection over union (IoU).
Additionally, to assess the proportion of training data points that generate workflows yielding correct results and are suitable for training, we use the term ``data utilization."

\textbf{Tool Library in Different Task Environment.}\label{subsection:toollibrary}
We follow previous works~\cite{khan2024visrep, ke2024hydra, suris2023vipergpt, gupta2023visprog, you2023idealgpt} on evaluation protocol using task-specific tool libraries for each task. These tool libraries might include BLIP2~\cite{li2023blip2}, LLaVA-1.5~\cite{liu2024improvedllava15}, GroundingDINO~\cite{liu2023groundingdino}, which are integrated into all compositional visual reasoning methods for fair comparison. Our framework is also a plug-and-play system, meaning that the performance can be enhanced by incorporating more advanced perception tools. For details on tool utilization methods, please refer to the appendix.

\begin{table*}[ht]
\centering
\caption{Comparison of average performance across six datasets, each corresponding to a distinct task. 
$\# Shots$ means the number of provided in-context learning examples. 
We use $*$ to highlight the second high score. For all E2E methods, we present their results \textcolor{lightgray}{in grey} as they serve as reference points but are not compositional visual reasoning methods and are not intended as direct comparison targets.}
\vspace{-0.5em}
\label{tab:main_table}
\scalebox{0.75}{\begin{tabular}{cccccccccc}
\toprule
      & & & VCR & EKVQA & VLCU & VASA & GD & CCQ\\ 
\midrule
Type & Method & $\# Shots$ & GQA  & OKVQA  & SugarCREPE & NaturalBench & RefCOCO & TallyQA \\
\midrule
\multirow{4}{*}{\textcolor{lightgray}{E2E}} & \textcolor{lightgray}{BLIP2~\cite{li2023blip2}} &  - & \textcolor{lightgray}{45.5} & \textcolor{lightgray}{31.4} & \textcolor{lightgray}{53.0} & \textcolor{lightgray}{2.9} & \textcolor{lightgray}{-} & \textcolor{lightgray}{49.0}\\
& \textcolor{lightgray}{LLaVa-1.5~\cite{liu2024improvedllava15}} & -  & \textcolor{lightgray}{62.1} & \textcolor{lightgray}{60.6} & \textcolor{lightgray}{52.7} & \textcolor{lightgray}{12.0} & \textcolor{lightgray}{34.9} & \textcolor{lightgray}{68.0}\\

&\textcolor{lightgray}{GPT-4o~\cite{hurst2024gpt4o}} & - & \textcolor{lightgray}{58.5} & \textcolor{lightgray}{33.4} & \textcolor{lightgray}{62.5} & \textcolor{lightgray}{16.8} & \textcolor{lightgray}{30.5} & \textcolor{lightgray}{76.4}\\

&\textcolor{lightgray}{YOLO-world~\cite{cheng2024yoloworld}} & - & \textcolor{lightgray}{-}& \textcolor{lightgray}{-} & \textcolor{lightgray}{-} & \textcolor{lightgray}{-} & \textcolor{lightgray}{36.6} &  \textcolor{lightgray}{-} \\

&\textcolor{lightgray}{GroundingDINO~\cite{liu2023groundingdino}} & - & \textcolor{lightgray}{-}& \textcolor{lightgray}{-} & \textcolor{lightgray}{-} & \textcolor{lightgray}{-} & \textcolor{lightgray}{80.5} &  \textcolor{lightgray}{-} \\

\midrule
\multirow{6}{*}{Compositional} &

Frozen few-shot & 10 & $48.8\pm2.2$ & $56.2\pm0.5$ & $56.3\pm0.9$ & $4.8\pm1.4$ & $63.0\pm1.8^*$ & $60.2\pm0.5^*$\\
&VisRep~\cite{khan2024visrep} & 10 & $51.4\pm1.0$ & $46.7\pm2.0$ & $58.2\pm1.6^*$ & $12.3\pm1.2^*$ & $55.2\pm0.8$ & $47.9\pm1.2$\\
& HYDRA~\cite{ke2024hydra} & 10 & $62.3\pm1.2^*$ & $60.6\pm1.4^*$ & $55.5\pm2.0$ & $12.3\pm1.9$ & $60.4\pm0.6$ & $57.2\pm2.3$\\
\addlinespace
\cline{2-9}
\addlinespace
& \textbf{DWIM} (Ours) & 0 & $\mathbf{69.3\pm1.0}$ & $\mathbf{62.8\pm1.2}$ & $\mathbf{74.6\pm1.3}$ & $\mathbf{13.6\pm1.4}$ & $\mathbf{82.7\pm0.9}$ & $\mathbf{72.0\pm1.3}$\\

& \textbf{\textcolor{orange}{$\Delta$(abs)} / \textcolor{blue}{$\Delta$(\%)}} & 

-
& $\uparrow$ \textbf{\textcolor{orange}{7.0} / \textcolor{blue}{11.2}} & $\uparrow$ \textbf{\textcolor{orange}{2.2} / \textcolor{blue}{3.6}}& $\uparrow$ \textbf{\textcolor{orange}{16.4} / \textcolor{blue}{28.2} }& $\uparrow$ \textbf{\textcolor{orange}{1.3} / \textcolor{blue}{10.6}} & $\uparrow$ \textbf{\textcolor{orange}{19.7} / \textcolor{blue}{31.3}}& $\uparrow$ \textbf{\textcolor{orange}{11.8} / \textcolor{blue}{19.6}}\\

\bottomrule
\end{tabular}}
\vspace{-0.5em}
\end{table*}

\begin{table*}[tb]
\centering
\caption{Cross-dataset Generalization Ability Study. We use the checkpoint trained on the GQA training set to test models on other datasets. Results for all E2E methods and frozen agentic LLMs are shown \textcolor{lightgray}{in grey} as reference points; these models are neither compositional nor trained and are not intended for direct comparison. $\Delta$(abs) means absolute difference and $\Delta$(\%) means relative difference.}
\vspace{-1em}
\label{tab:generalization_table}
\scalebox{0.65}{\begin{tabular}{cccccccc}
\toprule

Type & Method  & $\# Shots$ & OKVQA (\%) & A-OKVQA (\%) &  SugarCREPE (\%) & Winoground (\%)  \\
\midrule
\multirow{4}{*}{\textcolor{lightgray}{E2E}} & \textcolor{lightgray}{BLIP2~\cite{li2023blip2}} & \textcolor{lightgray}{-} & \textcolor{lightgray}{31.4} & \textcolor{lightgray}{41.0} & \textcolor{lightgray}{53.0} & \textcolor{lightgray}{49.9}  \\

&\textcolor{lightgray}{LLaVa-1.5~\cite{liu2024improvedllava15}} & \textcolor{lightgray}{-} & \textcolor{lightgray}{60.6} & \textcolor{lightgray}{68.7} & \textcolor{lightgray}{52.7} & \textcolor{lightgray}{49.9} \\

&\textcolor{lightgray}{GPT-4o~\cite{hurst2024gpt4o}} & \textcolor{lightgray}{-} & \textcolor{lightgray}{33.4} & \textcolor{lightgray}{63.2} & \textcolor{lightgray}{62.5} & \textcolor{lightgray}{65.6} \\

\midrule 
\multirow{4}{*}{Compositional} 

& VisRep~\cite{khan2024visrep} & 10 & $32.4\pm2.0$ & $63.5\pm1.6$ & $58.2\pm1.9$ & $49.3\pm0.4$ \\
& HYDRA~\cite{ke2024hydra} & 10 & $59.3\pm1.0$ & $62.1\pm0.5$ & $54.3\pm1.0$ & $50.1\pm1.2$ \\
 
\addlinespace
\cline{2-7}
\addlinespace
&\textbf{DWIM} (Ours on GQA) &0 & $\mathbf{60.8\pm0.8}$ & $\mathbf{69.8\pm1.3}$ & $\mathbf{62.4\pm1.2}$ & $\mathbf{57.4\pm0.1}$ \\

&\textbf{\textcolor{orange}{$\Delta$(abs)} / \textcolor{blue}{$\Delta$(\%)}} & $-$
& $\uparrow$ \textbf{\textcolor{orange}{1.5} / \textcolor{blue}{2.5}} & $\uparrow$ \textbf{\textcolor{orange}{6.3} / \textcolor{blue}{9.9}} & $\uparrow$ \textbf{\textcolor{orange}{4.2} / \textcolor{blue}{7.2}} & $\uparrow$ \textbf{\textcolor{orange}{7.3} / \textcolor{blue}{14.6}}\\

\bottomrule
\end{tabular}}
\vspace{-1em}
\end{table*}

\subsection{Quantitative Analysis}\label{sec:quantitative}

\textbf{Method Comparison Across Different Tasks:}
We compare our method with recent state-of-the-art (SOTA) compositional visual reasoning methods, including VisRep~\cite{khan2024visrep} and HYDRA~\cite{ke2024hydra}, as shown in Table~\ref{tab:main_table}. 
\textit{Frozen few-shot} refers to the method that uses a frozen LLaMa 3.1 within our designed inference framework. We report the average ACC/IoU score over three runs along with the standard deviation to ensure the reliability and consistency of our results. 
Both VisRep~\cite{khan2024visrep}, HYDRA~\cite{ke2024hydra} and \textit{Frozen few-shot} are evaluated with 10-shot in-context learning examples, as they rely heavily on in-context examples. In contrast, our method is evaluated without any in-context examples during the evaluation process because of our methods advantage. 
Additionally, we provide results for several end-to-end methods within the E2E block as reference points in Table~\ref{tab:main_table}, including the visual perception models used in all compositional approaches: \textbf{BLIP2}~\cite{li2023blip2}, \textbf{LLaVA-1.5}~\cite{liu2024improvedllava15} and \textbf{GroundingDINO}~\cite{liu2023groundingdino}.

Our method demonstrates SOTA performance across all tasks, achieving an average absolute improvement of over $9.73\%$ compared to existing compositional approaches, and outperforming all utilized tools.

\textbf{Generalization Ability Comparison:}
To evaluate generalization, we assess the cross-dataset performance of compositional approaches. Specifically, we use the checkpoint trained on GQA~\cite{hudson2019gqa} to test models on datasets with the same inference setup (i.e., OKVQA~\cite{marino2019okvqa}, A-OKVQA~\cite{schwenk2022aokvqa}, SugarCREPE~\cite{hsieh2024sugarcrepe}, and Winoground~\cite{thrush2022winoground}), ensuring a fair comparison.

As shown in Table~\ref{tab:generalization_table}, our method demonstrates strong generalization abilities, even without relying on in-context learning examples. 
Our method achieves an approximately 4.8\% improvement compared to HYDRA~\cite{ke2024hydra} and VisRep~\cite{khan2024visrep}, highlighting its effectiveness and strong zero-shot performance. Two E2E models, including LLaVa-1.5~\cite{liu2024improvedllava15} and BLIP2~\cite{li2023blip2}, trained on datasets like GQA and OKVQA, serve as reference points. Our method achieves superior zero-shot cross-dataset performance, outperforming baselines and demonstrating capabilities comparable to GPT-4o.

\begin{figure}[t]
\begin{center}
  \includegraphics[width=0.8\linewidth]{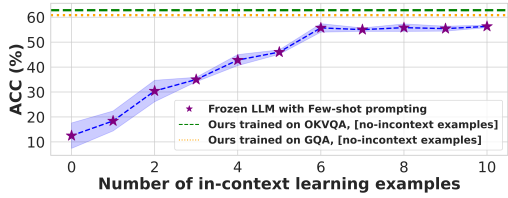} 
\end{center}
\vspace{-1.5em}
\caption{Frozen LLM with in-context learning example \emph{v.s.} 0-shot trained LLM performance on OKVQA dataset. }
\label{fig:shot_diff}

\end{figure}

\textbf{Frozen LLMs encounter performance bottlenecks} even as the number of in-context learning examples increases, compared to DWIM.
We conducted experiments to demonstrate the robustness of our method compared to frozen LLMs with few-shot prompting.
As illustrated in Figure~\ref{fig:shot_diff}, frozen agentic LLMs do not have ability to understand provided tool and do not inherit with problem solving logic ability and only get around 12\% ACC on OKVQA dataset. As the number of in-context learning examples increases, the performance of frozen LLMs improves but quickly reaches a saturation point. This suggests that further gains from extensive human prompt engineering may be marginal, as scaling up the number of examples yields diminishing returns. While examples can teach the LLM how to use tools, they cannot cover all possible situations where tools may fail. Overall, our method outperforms existing approaches, achieving significantly better performance with a small training dataset.

\textbf{Dependence of Different Compositional Methods on Task-Specific Tool Libraries.} All previous experiments follow the evaluation protocol, using task-specific tool libraries for each task. In this experiment, we train and evaluate all compositional methods using a complete tool library instead of a task-specific one on two tasks to examine their dependence on designed tool libraries. Results in Table~\ref{tab_all_tool} show that using a complete library decreases performance across all methods. This decline stems from the challenge of selecting optimal tools from a larger selection space, worsened by frozen LLMs' limited tool awareness. However, the greater performance disparity highlights DWIM's effectiveness and reduced reliance on manual adjustments.

\begin{table}
\vspace{-2mm}
\caption{Comparison on average performance, TS: task-specific}
\vspace{-3mm}
\label{tab_all_tool}
\centering
\scalebox{0.63}{
\begin{tabular}{ccccccc}
\toprule[0.3mm]
\rowcolor[gray]{0.95} Method & $\# Shots$ & \multicolumn{2}{c}{GQA (\%)} & \multicolumn{2}{c}{SugarCREPE (\%)} \\
\rowcolor[gray]{0.95}& & TS & Complete & TS & Complete \\
\cline{3-4}\cline{5-6}
Frozen few-shot & 10 & $48.8\pm2.2$ & $40.5 \pm1.4$ & $56.3\pm0.9$ &  $40.9\pm0.4$\\
VisRep & 10 & $51.4\pm1.0$  & $41.7\pm3.6$ & $58.2\pm1.6$  & $46.4\pm1.8 $\\
HYDRA & 10 & $62.3\pm1.2$  & $53.3\pm2.6$ & $55.5\pm2.0$  & $43.6\pm1.9$\\

\cline{1-6}

DWIM (Ours) & 0 & $\mathbf{69.3\pm1.0}$ & $\mathbf{67.1\pm0.2}$ & $\mathbf{74.6\pm1.3}$ & $\mathbf{63.0\pm1.4}$\\
\textbf{\textcolor{orange}{$\Delta$(abs)} / \textcolor{blue}{$\Delta$(\%)}} & -
& $\uparrow$ \textbf{\textcolor{orange}{7.0} / \textcolor{blue}{11.2}} & $\uparrow$ \textbf{\textcolor{orange}{13.8} / \textcolor{blue}{25.9}} & $\uparrow$ \textbf{\textcolor{orange}{16.4} / \textcolor{blue}{28.2}} & $\uparrow$ \textbf{\textcolor{orange}{16.6} / \textcolor{blue}{35.8}}\\

\bottomrule[0.3mm]
\end{tabular}}
\vspace{-2em}
\end{table}

\textbf{Comparison of Tool-Use Efficiency Across Methods:}
We quantified the average tool utilization per successful sample to assess tool awareness from an efficiency perspective. Each tool use was counted as a single instance, yielding the following averages: Frozen few-shot: 2.87, HYDRA: 3.26, VisRep: 2.49, and DWIM: \textbf{1.74}. Lower values indicate better performance, highlighting DWIM's superior tool awareness compared to frozen LLMs. Further analysis of tool awareness is provided in the supplementary material.

\subsection{Qualitative Analysis}
We provide two qualitative analyses in the Appendix: one assessing DWIM's performance and another comparing training workflow generation.

\subsection{Ablation Studies}

In this section, we provide ablation studies on the key techniques proposed in \ours{} to demonstrate their contributions to the final results.  

\textbf{Efficient Training Workflow Generation and Collection.}
In this study, we investigate the impact of discrepancy-aware training workflow generation on data utilization. Data utilization refers to the proportion of training data points that can generate workflows yielding correct results and are suitable for training. 
As shown in Table~\ref{subtab:ablaton_trajectory_generation}, we compare our method with the standard generation approach on the GQA and SugarCREPE datasets.
The collected workflows are then used to fine-tune LLaMa3.1-8B, and performance is evaluated using Instruct Masking without in-context learning examples. On GQA, our approach improves data utilization by 17.4\%, enabling more valid workflows and achieving an 11.3\% absolute performance gain over the standard method. Similarly, on SugarCREPE, it increases data utilization by 17.1\% and boosts absolute performance by 13.9\%. This comparison highlights the effectiveness of our method in maximizing data utility and achieving optimal outcomes in visual reasoning tasks.

\begin{table}[tb]
\centering
\caption{Ablation Study: Effectiveness of the proposed training workflow generation. W/O denotes the use of the ``standard" (as described in Sec \ref{subsection:DWIM} Eq.\ref{qe:standardmethod}) method, while W indicates the use of the discrepancy-aware workflow generation strategy.}
\vspace{-0.5em}
\label{subtab:ablaton_trajectory_generation}
\scalebox{0.6}{\begin{tabular}{ccccc}
\toprule
Dataset & Type & Generator & Data Utilization (\%) & ACC (\%) \\ 
\midrule
\multirow{4}{*}{GQA} 
    & W/O & LLaMa3.1(8B) & 48.2 & 57.9 \\
    & W/O & GPT4O & 50.9 & 58.0  \\
    & W & LLaMa3.1(8B) & \textbf{68.3} & \textbf{69.3}  \\

 & & \textbf{\textcolor{orange}{$\Delta$(abs)} / \textcolor{blue}{$\Delta$(\%)}} & $\uparrow$ \textbf{\textcolor{orange}{17.4} / \textcolor{blue}{34.2}} & $\uparrow$ \textbf{\textcolor{orange}{11.3} / \textcolor{blue}{19.5} }\\
\midrule
\multirow{4}{*}{SugarCREPE} 
    & W/O & LLaMa3.1(8B) & 56.5 & 60.0 \\
    & W/O & GPT4O & 59.3 & 60.7 \\
    & W & LLaMa3.1(8B) & \textbf{76.4} & \textbf{74.6} \\

& & \textbf{\textcolor{orange}{$\Delta$(abs)} / \textcolor{blue}{$\Delta$(\%)}} & $\uparrow$ \textbf{\textcolor{orange}{17.1} / \textcolor{blue}{28.8} } & $\uparrow$ \textbf{\textcolor{orange}{13.9} / \textcolor{blue}{22.9} } \\

\bottomrule
\end{tabular}}
\vspace{-1.5em}
\end{table}

\textbf{Instruct-Masking Fine-tuning.} 
In this ablation study, we examine the effectiveness of the instruct-masking fine-tuning method compared to SFT on two tasks using GQA and SugarCREPE datasets as shown in Table~\ref{subtab:ablaton_masking_training}. Leveraging workflows generated via the standard method, instruct-masking yields noticeable performance gains, with a 4.3\% absolute performance improvement on the GQA dataset and a 2.9\% absolute improvement on SugarCREPE. However, when paired with our discrepancy-aware training workflow generation and collection strategy, the absolute performance gains become even more pronounced, reaching 14.5\% on GQA and 16.3\% on SugarCREPE.
Furthermore, instruct-masking exhibits a clear advantage over both \textit{Random-Masking} and \textit{Masking-W-Rethink}. Random-Masking indiscriminately masks actions rather than selectively targeting correct actions within the workflow, while Masking-W-Rethink masks both effective actions and ineffective actions that trigger ``Rethink." These experimental results collectively demonstrate the effectiveness of instruct-masking in training on noisy workflows, leading to improved efficiency and accuracy in the training process.

\begin{table}[tb]
\centering
\caption{Ablation Study: Instruct-Masking Fine-tuning on GQA and SugarCREPE. ``Random-Masking'' refers to the process of randomly masking any action within a workflow and is shown \textcolor{lightgray}{in grey} because it is neither an existing method nor our proposed method, but is included for the ablation study.}
\vspace{-1em}
\label{subtab:ablaton_masking_training}
\scalebox{0.6}{%
\begin{tabular}{cccccc}
\toprule
Fine-tune & Method & Model & GQA & SugarCREPE\\ 
\midrule
\addlinespace

Frozen & 10-shot prompting & LLaMa3.1 (8B) & 48.8 & 56.3\\
Frozen & 10-shot prompting & GPT4o & 51.0 & 57.7\\

\midrule

SFT & Standard & LLaMa3.1 (8B) & 53.6 & 57.1\\

Instruct-Masking & Standard & LLaMa3.1 (8B) & 57.9 & 60.0\\

& &  \textbf{\textcolor{orange}{$\Delta$(abs)} / \textcolor{blue}{$\Delta$(\%)}} & $\uparrow$ \textbf{\textcolor{orange}{4.3} / \textcolor{blue}{8.0}} & $\uparrow$ \textbf{\textcolor{orange}{2.9} / \textcolor{blue}{5.1}}\\

\midrule

SFT & Discrepancy-aware & LLaMa3.1 (8B) & 54.8 & 58.3\\

\textcolor{lightgray}{Random-Masking} & \textcolor{lightgray}{Discrepancy-aware} & \textcolor{lightgray}{LLaMa3.1 (8B)} & \textcolor{lightgray}{65.1} & \textcolor{lightgray}{63.5}\\

\textcolor{lightgray}{Masking-W-Rethink} & \textcolor{lightgray}{Discrepancy-aware} & \textcolor{lightgray}{LLaMa3.1 (8B)} & \textcolor{lightgray}{68.0} & \textcolor{lightgray}{70.7} \\

Instruct-Masking & Discrepancy-aware & LLaMa3.1 (8B) & 69.3 & 74.6\\

& &  \textbf{\textcolor{orange}{$\Delta$(abs)} / \textcolor{blue}{$\Delta$(\%)}} & $\uparrow$ \textbf{\textcolor{orange}{14.5} / \textcolor{blue}{26.5}} & $\uparrow$ \textbf{\textcolor{orange}{16.3} / \textcolor{blue}{28.0}}\\

\bottomrule
\end{tabular}}
\vspace{-0.5em}
\end{table}

\begin{table}[t]
\centering
\caption{LLM Backbone Experiments on VCR Task.}
\vspace{-1em}
\label{tab:backbone_table}
\scalebox{0.6}{\begin{tabular}{ccccc}
\midrule
\rowcolor[gray]{0.95}Method & LLaMa-3.1-8B (\%) & Mistral-v0.3-7B (\%) & Mistral-v0.2-7B (\%)\\
\midrule
Frozen few-shot & 48.8 & 52.0 & 45.1  \\

\textbf{DWIM} (Ours) & 69.3 & 67.5 & 63.7\\

 \textbf{\textcolor{orange}{$\Delta$(abs)} / \textcolor{blue}{$\Delta$(\%)}} 
& $\uparrow$ \textbf{\textcolor{orange}{20.5} / \textcolor{blue}{42.0}} & $\uparrow$ \textbf{\textcolor{orange}{15.5} / \textcolor{blue}{29.8}} & $\uparrow$ \textbf{\textcolor{orange}{18.6} / \textcolor{blue}{41.2}} \\

\midrule
\rowcolor[gray]{0.95}Method & Qwen2.5-1.5B (\%) &  Qwen2.5-3B (\%) &  Qwen2.5-7B (\%)\\
\midrule
Frozen few-shot      & 18.7 & 41.4 & 53.4 \\

\textbf{DWIM} (Ours) & 48.2 & 63.0 & 65.8\\

 \textbf{\textcolor{orange}{$\Delta$(abs)} / \textcolor{blue}{$\Delta$(\%)}} 
& $\uparrow$ \textbf{\textcolor{orange}{29.5} / \textcolor{blue}{157.6}} & $\uparrow$ \textbf{\textcolor{orange}{21.6} / \textcolor{blue}{52.2}} & $\uparrow$ \textbf{\textcolor{orange}{12.4} / \textcolor{blue}{23.2}} \\
\bottomrule
\end{tabular}}
\vspace{-1.5em}
\end{table}

\textbf{Experiment Using Various LLM Backbones for DWIM.}
All previous results show that DWIM significantly enhances LLaMa-3.1's capabilities. To evaluate its generality, we apply DWIM to various open-source models (e.g., LLaMa-3.1-8B~\cite{llama3120243}, Mistral-v0.2/0.3-7B~\cite{jiang2023mistral}, and three Qwen2.5 variants~\cite{qwen2025qwen25technicalreport}), and compare their performance to frozen LLMs with 10-shot in-context learning. As shown in Table~\ref{tab:backbone_table}, DWIM improves absolute performance by over 12.4\% across all models, demonstrating its effectiveness.

%% file: sec/5_conclusion.tex
\section{Conclusion}
\label{sec:conclusion}
This paper introduces an efficient discrepancy-aware training workflow generation and instruct-masking fine-tuning method for tool-aware visual reasoning. These innovations address the challenges of efficiently generating workflow data within environments with potential incorrect feedback and fine-tuning on noisy workflows. DWIM enhances LLM tool awareness, enabling more accurate and efficient tool usage for complex visual reasoning tasks. Extensive experiments on public benchmarks demonstrate that our method achieves state-of-the-art performance, offering a more generalizable and robust solution for compositional visual reasoning tasks.



%% file: sec/X_suppl.tex
\clearpage
\setcounter{page}{1}
\maketitlesupplementary

\section{Auto-Exploring Agentic Framework}

\begin{figure}[t]
\begin{center}
  \includegraphics[width=0.95\linewidth]{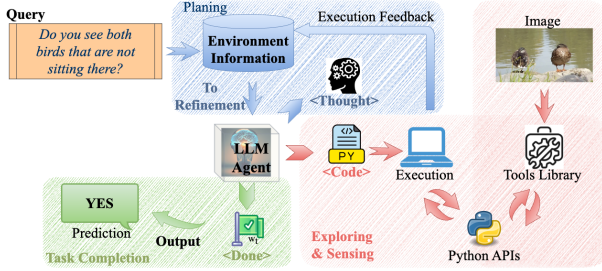} 
\end{center}
\vspace{-2em}
\caption{Auto-Exploring Agentic Framework. 
The LLM agent generates \text{\textlangle Code\textrangle} for execution, \text{\textlangle Thought\textrangle} for reasoning, or \text{\textlangle Done\textrangle} to complete the task. 
It dynamically generates or refines actions while storing environmental information for incremental reasoning.}
\vspace{-2em}
\label{fig:framework}
\end{figure}

Our framework dynamically generates \text{\textlangle Code\textrangle}, \text{\textlangle Thought\textrangle}, or \text{\textlangle Done\textrangle} without a fixed pattern (e.g., Code follows Thought in CodeAct~\cite{pmlrcodeact}, or Act follows Thought in ReAct~\cite{yao2022react}). The LLM generates \text{\textlangle Code\textrangle}  for all execution steps involving tool usage. If reasoning is required or an ineffective action is detected by the discrepancy-aware recognition step, the model outputs the corresponding information in \text{\textlangle Thought\textrangle}. This flexibility makes our model inherently dynamic.

\section{DWIM Qualitative Analysis}

\begin{figure}
\begin{center}
  \includegraphics[width=0.9\linewidth]{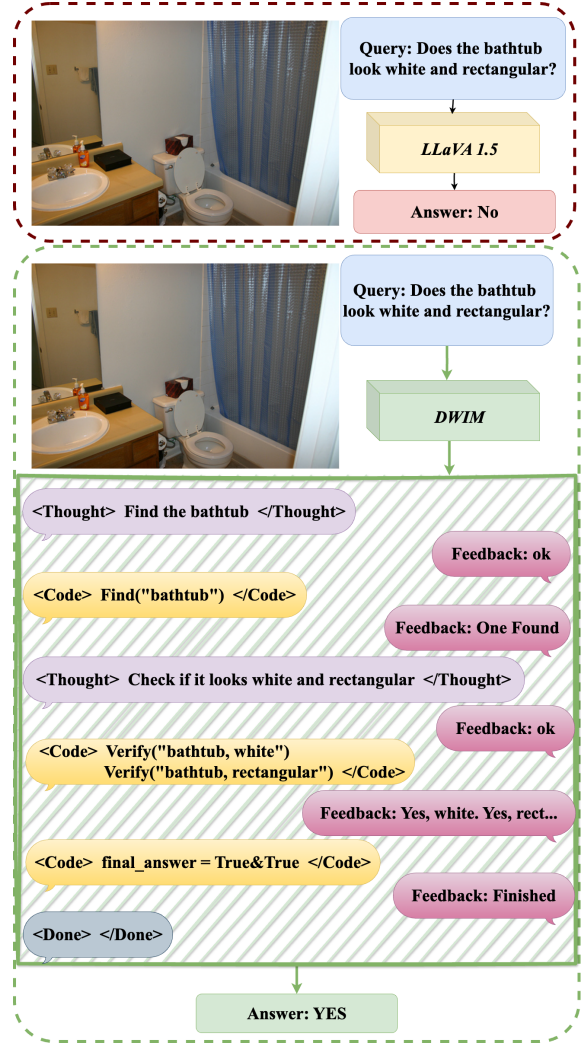} 
\end{center}
\vspace{-1.5em}
\caption{DWIM Qualitative Result Example}
\label{fig:qualitative}
\vspace{-2em}
\end{figure}

In this section, we provide a qualitative analysis showcasing the output of each step in DWIM, as illustrated in Figure~\ref{fig:qualitative}. The input image, located at the top-left of each bounding box, and the query are displayed in the light blue box.
The \textcolor{darkpurple}{purple box displays the \text{\textlangle Thought\textrangle}-action}, the \textcolor{lightlydarkyellow}{yellow box shows the \text{\textlangle Code\textrangle}-action}, and the \textcolor{darkpink}{pink box presents the environment feedback}, for each turn respectively. In many cases, LLaVa-1.5~\cite{liu2024improvedllava15}, one of the tools in our tool library, fails to answer the question. In comparison, by leveraging the tool-awareness ability, DWIM provides correct answers by utilizing tools better suited for the question.

\section{``Standard'' \emph{v.s.} ``Discrepancy-aware'' Training Workflow Generation}
\begin{figure*}
\begin{center}
  \includegraphics[width=0.9\linewidth]{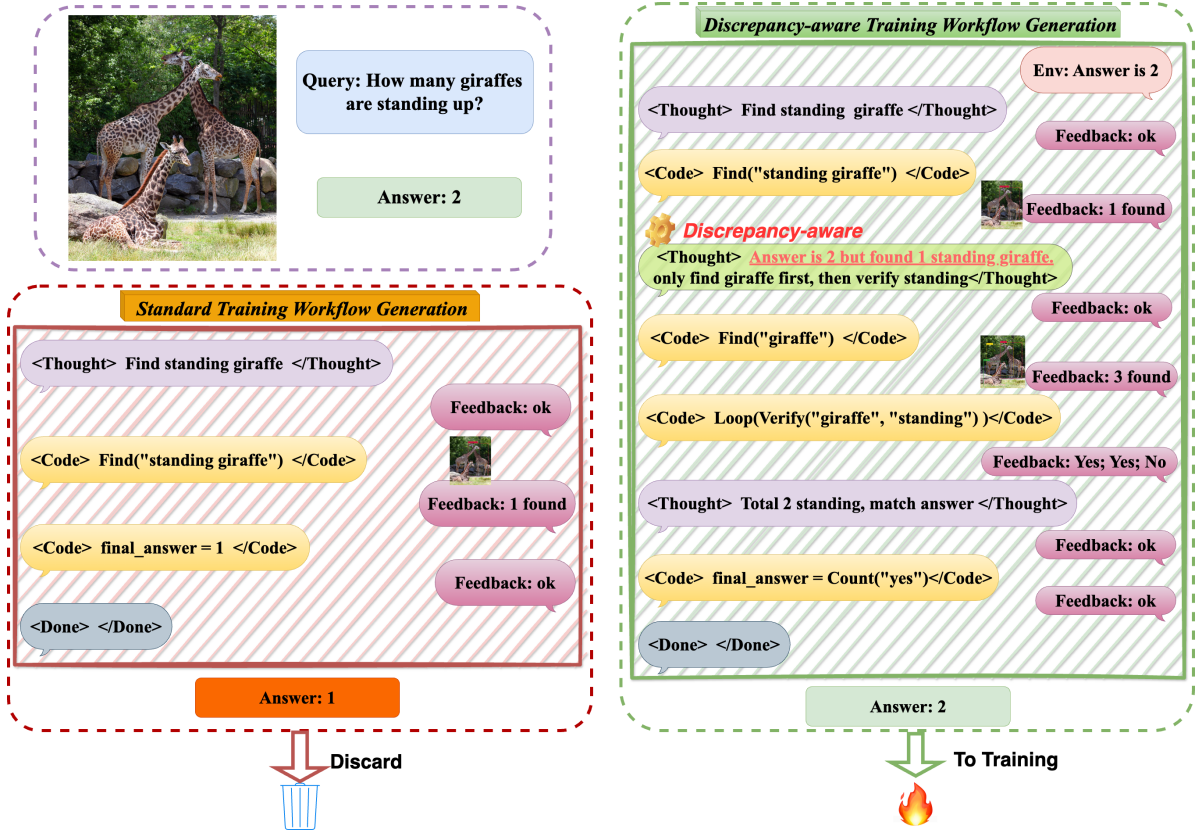} 
\end{center}
\vspace{-1.5em}
\caption{``Standard'' \emph{v.s.} ``Discrepancy-aware training'' Training Workflow Generation}
\label{fig:workflow_generation_diff}
\end{figure*}

In this section, we provide a qualitative analysis of the differences between standard training workflow generation and discrepancy-aware training workflow generation, as shown in Figure~\ref{fig:workflow_generation_diff}. Using the standard method, the model assumes environmental feedback is always correct under the same auto-exploring framework. As a result, the standard method does not check for discrepancies between feedback information and the answer, leading to failed workflows due to tool errors and preventing the generation of a viable workflow for that training data point. Consequently, a large portion of training data lacks correct workflows that yield the right answers and is discarded, resulting in high data waste. In contrast, discrepancy-aware training workflow generation accounts for discrepancies between each feedback step and the answer to ensure that actions remain valid while continuously refining workflows to reach the correct final answer. This discrepancy-aware step also enables the model to recognize when tools provide incorrect information, which is crucial for training agent tool awareness.

\section{Analysis of Action Flagging}\label{lab_action_flagging}
In DWIM, flagging action effectiveness based on both LLM assessments and environment feedback is a prerequisite for instruct-masking. The LLM identifies discrepancies between feedback and the expected answer by generating descriptive sentences (\eg, $\omega_{\text{Rethink}}$).  We assess action effectiveness using both the content of these sentences and the corresponding feedback. To support this, we employ a rule-based method that flags ineffective actions based on discrepancy-aware recognition and environmental feedback. These flagged actions are excluded from masking, preventing the model from learning from mistakes. However, LLM assessment output may not fully adhere to the output template when recognizing ineffective actions in a workflow due to its complexity, which involves natural language, code, and intricate environment feedback, potentially leading to misflagging.

\begin{figure*}
\begin{center}
  \includegraphics[width=0.7\linewidth]{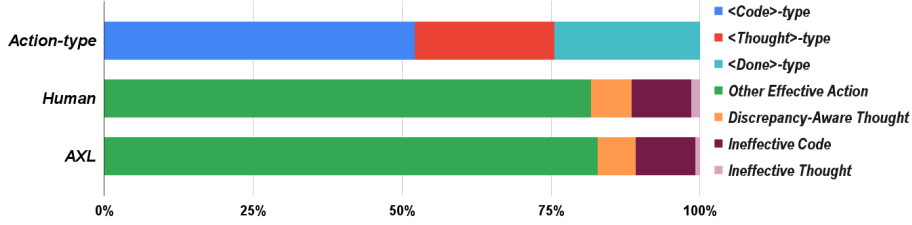} 
\end{center}
\vspace{-1em}
\caption{DWIM and Human Flagging of Ineffective Actions on Collected Workflows. DWIM's flagging results are close to those of human evaluators; however, there is still room for improvement, particularly in flagging actions that trigger ``rethink.''}
\label{fig:effectiveness_analysis}
\end{figure*}

To evaluate our proposed flagging method, we conduct a human evaluation to assess the effectiveness of flagging in $100$ workflow samples generated using the discrepancy-aware training workflow generation method from the GQA training set, comparing the results with our rule-based approach.
In these 100 workflow samples, $52.1\%$ are \text{\textlangle Code\textrangle}-actions, $23.5\%$ are \text{\textlangle Thought\textrangle}-actions, and $24.4\%$ are \text{\textlangle Done\textrangle}-actions.

In DWIM, any \text{\textlangle Code\textrangle}-action flagged by environment feedback as ``Traceback'' is flagged as ineffective. Similarly, a action preceding a \text{\textlangle Thought\textrangle}-action (\eg, $\omega_{\text{Rethink}}$) with the context ``however'' or ``rethink'' is also considered ineffective. An action that is logically correct but produces an incorrect result will trigger a discrepancy-aware \text{\textlangle Thought\textrangle}-action.
A total of $41$ ineffective \text{\textlangle Code\textrangle}-actions, $3$ actions preceding a \text{\textlangle Thought\textrangle}-action with the context ``rethink,'' and $26$ discrepancy-aware \text{\textlangle Thought\textrangle}-action were flagged as ineffective. In the human evaluation, we used the majority vote from three evaluators and obtained the same results for normal ineffective \text{\textlangle Code\textrangle}-actions. Additionally, $3$ more actions triggering ``rethink'' or ``replan'' \text{\textlangle Thought\textrangle}-actions and 2 additional discrepancy-aware \text{\textlangle Thought\textrangle} actions were identified.

\begin{table*}[t]
\centering
\caption{Tools' Functionality}
\label{tab:tool_details}
\scalebox{0.7}{\begin{tabular}{cccccc}
\toprule
\rowcolor[gray]{0.95} Tools & \multicolumn{4}{c}{Model Name} &Description\\
\cline{2-5}
\rowcolor[gray]{0.95}  & LLaVa-1.5-7B & BLIP2-Flan-T5-XXL & GPT-4o-2024-05-13 & GroundingDINO-Base &  \\ 
\midrule
Detector                       &  &  &  & \checkmark & Detect Object \\
Check Existence                 &  &  &  & \checkmark & Check Object Existence\\
Simple Query Answer             & \checkmark & \checkmark &  & & Answering Simple Questions with a Word or Phrase\\
Complex Query Answer            &  \checkmark   &  &  &  & Answering Complex Questions with a Sentence \\
Captioning                      & \checkmark &  &  & & Get Image Caption\\
Acquiring External Knowledge    &   &  &   \checkmark  & & Acquire External Knowledge \\
Boolean to Yes/No               &  &    &  & & Convert True/False to Yes/No\\
Image Crop                      &  &  &  & & Crop Images Based on Provided Coordinates\\
Property Matching               & \checkmark &  &  &  & Identify the Best-Matching Visual Property \\
Verify Property                 & \checkmark &  &  &  &Verify Visual Property\\
\bottomrule
\end{tabular}}
\end{table*}

\begin{table}[tb]
\centering
\caption{Task-specific Tool Library.}
\label{tab:tools}
\scalebox{0.65}{\begin{tabular}{ccccccc}
\toprule
\rowcolor[gray]{0.95}Tools  & \multicolumn{6}{c}{Tasks} \\
\cline{2-7}
\rowcolor[gray]{0.95} & VCR & EKVQA & VLCU & VASA & GD & CCQ \\ 
\midrule
Detector      & \checkmark & \checkmark & \checkmark & \checkmark  & \checkmark & \checkmark\\

Check Existence                 & \checkmark & \checkmark & \checkmark & \checkmark & \checkmark & \checkmark\\
Simple Query Answer             & \checkmark & \checkmark & \checkmark & \checkmark & \checkmark & \checkmark\\
Complex Query Answer            &            & \checkmark &            &           \\
Captioning                      & \checkmark & \checkmark & \checkmark & \checkmark & \checkmark & \checkmark\\
Acquiring External knowledge    &            & \checkmark &            &           \\
Boolean to Yes/No               & \checkmark &            & \checkmark & \checkmark\\
Image Crop                      & \checkmark & \checkmark & \checkmark & \checkmark & \checkmark & \checkmark\\
Property Matching               & \checkmark &            &            &           & \checkmark &\\
Verify Property                 & \checkmark & \checkmark & \checkmark & \checkmark & \checkmark & \checkmark\\
\bottomrule
\end{tabular}}
\end{table}

As illustrated in Figure~\ref{fig:effectiveness_analysis}, all normal ineffective \text{\textlangle Code\textrangle}-actions were flagged; however, only 50\% of ineffective \text{\textlangle Thought\textrangle}-action preceding a \text{\textlangle Thought\textrangle}-action with the context ``rethink'' were detected. Although such ineffective actions constitute only around $10\%$ of the total sample actions, it is crucial that they are not masked and are properly learned. The current rule-based flagging method is not entirely precise, particularly in complex contexts. In future work, we aim to develop an LLM-based flagger for more accurate flagging of ineffective actions by leveraging environment feedback and recognition results.

\section{Additional Ablation Study}
We conducted an experiment where the answer is given but discrepancies are not recognized (annotated as \textit{Given Y}) as shown in Table~\ref{subtab:additional_ab}. While \textit{Given Y} produces more workflows than the standard method, it does not outperform our approach. Moreover, a portion of its successful workflows result from directly copying the answer.
\begin{table}[ht]
\centering
\caption{Additional Ablation Study: Effect of \textit{Given Y} During Workflow Generation on GQA}
\label{subtab:additional_ab}
\scalebox{0.6}{%
\begin{tabular}{ccccc}
\toprule
\rowcolor[gray]{0.95}Fine-tune & Training Workflow Generation & Data Utilization (\%)  & GQA (\%) \\ 
\midrule
\addlinespace

SFT & Standard & 48.2 & 53.6 \\

Instruct-Masking & Standard & 48.2  & 57.9 \\

\midrule

SFT &  Given Y  & 60.3 & 54.3\\

Instruct-Masking &  Given Y  & 60.3 & 60.9\\

\midrule

SFT & Discrepancy-aware  & 68.3 & 54.8 \\

Random-Masking & Discrepancy-aware & 68.3  & 65.1 \\

Masking-W-Rethink & Discrepancy-aware & 68.3  & 68.0 \\

Instruct-Masking & Discrepancy-aware & \textbf{\textcolor{mypink}{68.3}} & \textbf{\textcolor{mypink}{69.3}} \\

\bottomrule
\end{tabular}}
\end{table}

\section{Additional Tool Awareness analysis}
We evaluate models' tool awareness based on overall performance and tool utilization efficiency, as described in Section~\ref{sec:experiment}. To further evaluate the improvement in tool awareness of DWIM compared to a frozen LLM, we conduct a human evaluation on 100 workflow samples from GQA evaluation results for each model. Specifically, we examine the proportion of generated workflows that should yield correct answers if the tools function accurately but fail in practice, as well as the proportion of workflows that are logically incorrect.

The evaluation results indicate that 71\% of DWIM-generated workflows and 47\% of frozen LLM-generated workflows produced correct answers. Additionally, 18\% and 28\% of workflows, respectively, should yield correct answers but failed due to tool errors. Furthermore, 7\% of DWIM-generated workflows and 23\% of frozen LLM-generated workflows were logically incorrect. Lastly, 4\% and 2\% of workflows, respectively, produced correct answers but were misclassified as incorrect due to evaluation metric errors.

Based on our investigation, we observe a significant improvement in overall performance after training, indicating the effectiveness of the generated workflows. Additionally, the average tool utilization per query decreases, suggesting improved efficiency. Moreover, DWIM has a 10\% lower failure rate than the frozen LLM in generating workflows that should produce correct answers but fail due to tool errors. This suggests that DWIM has a better understanding of each tool. Besides, DWIM is less likely to misuse tools when constructing workflows after training. Overall, these findings demonstrate that DWIM significantly enhances tool awareness.

\begin{figure*}[t]
\begin{center}
  \includegraphics[width=1\linewidth]{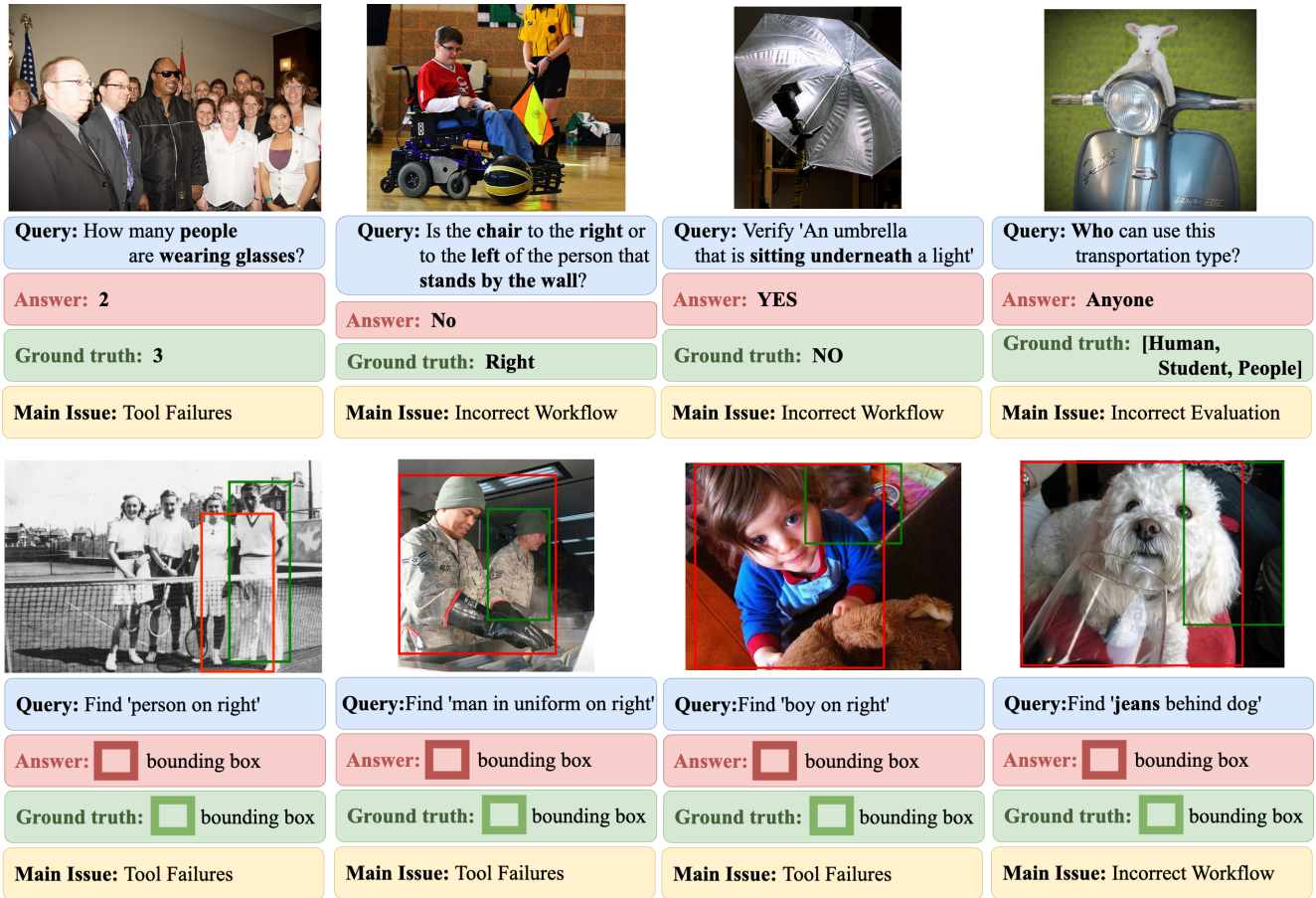} 
\end{center}
\caption{Failure Case Analysis. \textcolor{blue}{Queries are presented in blue boxes}, \textcolor{red}{DWIM's answers are displayed in red boxes}, and \textcolor{darkgreen}{ground truth labels are shown in green boxes}. Additionally, we provide the \textcolor{lightlydarkyellow}{main issues causing DWIM to fail in completing the task in yellow boxes}.}
\label{fig:failure_case}
\end{figure*}

\section{Tool Library and Functionality}
In this section, we introduce the details of the task-specific tool library (Table~\ref{tab:tool_details}), including the functionalities of each tool and their corresponding models. Table~\ref{tab:tools} provides a comprehensive overview of the tools included in the proposed tool library and their respective functionalities. The table is structured to showcase the capabilities of each tool across different models (LLaVA-1.5-7B~\cite{liu2024improvedllava15}, BLIP2-Flan-T5-XXL~\cite{li2023blip2}, GPT-4o~\cite{hurst2024gpt4o}, and GroundingDINO-Base~\cite{liu2023groundingdino}) and provides a brief description of their specific functionalities.
\begin{itemize}
    \item \textbf{Detector}: This functionality, supported by GroundingDINO, focuses on detecting objects within an image.
    \item \textbf{Check Existence}: GroundingDINO is also capable of checking the existence of specific objects within a given scene, contributing to basic visual verification tasks.
    \item \textbf{Simple Query Answer}: Both LLaVa-1.5 and BLIP2 excel in answering simple questions using a single word or phrase. This capability is valuable for tasks requiring concise and precise responses.
    \item \textbf{Complex Query Answer}: LLaVa-1.5 extends its capability to answering more complex questions, providing sentence-level responses that demand a deeper understanding of the image and associated context.
    \item \textbf{Captioning}: LLaVa-1.5 further supports image captioning, generating descriptive captions for input images to facilitate contextual interpretation.
    \item \textbf{Acquiring External Knowledge}: GPT-4o is the sole tool in this library designed to acquire external knowledge, which is essential for tasks that require external information beyond the given visual input.
    \item \textbf{Boolean to Yes/No}: This functionality would involve converting boolean values (True/False) into human-readable yes/no responses.
    \item \textbf{Image Crop}: This functionality is designed to crop images based on provided coordinates.
    \item \textbf{Property Matching}: It supports identifying the best-matching visual property among a set of options.
    \item \textbf{Verify Property}: It is capable of verifying visual properties.
    
\end{itemize}


\section{Failure Case Analysis}
While DWIM has achieved SoTA performance, there remains room for improvement in its design. In complex cases, as illustrated in Figure~\ref{fig:failure_case}, DWIM may fail due to errors made by the LLMs, resulting in incorrect workflows or workflows that are logically correct but fail due to tool errors. In future iterations, we aim to enhance the ability of agentic LLMs to automatically select and utilize tools for better decision-making.

Furthermore, we investigate the primary limitations of current frozen LLMs when presented with 10-shot examples. Through human investigation of workflows leading to incorrect answers provided by frozen LLMs, we identified the following common issues: \textbf{lack of reasoning ability to determine when to stop}, \textbf{lack of self-correction ability}, and \textbf{lack of tool awareness}, meaning the proposed methods are logically correct but practically flawed.

\section{Computational Costs}
Running on four RTX A6000 GPUs, the average inference and training time per sample (in seconds) is as follows: DWIM ($9.4$, $14.4$), HYDRA~\cite{ke2024hydra} ($3.6$, $28.8$), and VisRep~\cite{khan2024visrep} ($7.2$, $7.2$). HYDRA uses DQN for training, which is difficult to parallelize due to time constraints, and its official code does not support multi-GPU acceleration. Therefore, HYDRA training was conducted on a single RTX A6000 GPU.

To explore more computation information, we computed the average token count per sample for LLM of each method as shown in Table~\ref{tab:token_length}. Our method incurs slightly more computation than VisRep but achieves significantly better performance, while requiring far less than HYDRA (which uses GPT) and still outperforming it. 
\begin{table}[h]
\centering
\caption{Average Input and Output Token Counts of the LLM.}
\label{tab:token_length}
\scalebox{0.7}{\begin{tabular}{ccccc}
\midrule
\rowcolor[gray]{0.95} & DWIM (Ours) & VisRep (CVPR24) & HYDRA (ECCV24)\\
\midrule
Avg. Tokens (In+Out) & 5931.87 & 3520.44 & \textbf{9387.24}  \\

\bottomrule
\end{tabular}}

\end{table}

\section{Prompt Template}
In DWIM, the agentic LLM can autonomously explore the environment through three types of actions, as outlined in Section~\ref{sec:methodology}. In this section, we are providing both the prompt template for agent auto-exploration and the Python interface code enabling the agent's perception capabilities.

\begin{prompt}[{Auto-Exploring}]
\begin{lstlisting}[language=Python]
Your job is to write code to solve questions about images. You have access to the ImagePatch class above.
You will be able to interact with a Jupyter notebook. You have to carefully format your responses according to the following rules.

1. When you want to write code, you must use triple backticks inside a `<code>` tag. 
2. When you want to return text you must use the `<thought>` tag. Example: `<thought>I think this is the answer.</thought>`
3. When you are done, you must use the `<done>` tag with no content inside. Example: `<done></done>`
4. The response from the notebook will be enclosed inside a `<result>` tag. Example: `<result>2</result>`
5. The image will be loaded for you in a variable called `image`, the image detial captioning will be provided.
6. If you can directly answer the question using a single word or phrase, Your final answer should be stored in a variable called `final_answer`.
7. If you need more information, you can write code to get more information from image.
8. In each step, you can only use a _single_ action.
9. Take care to indent multi-line code carefully, and think step by step to solve the problem incrementally.
10. Answer the question using a single word or phrase and store the answer in `final_answer`, then exit the task with a `<done>` tag.
11. You must provide a solution, and please do not refuse to answer even if you are not completely sure.
12. If `final_answer` is `True` or `False`, please use `bool_to_yesno` to convert it to 'yes' or 'no'.
\end{lstlisting}
\end{prompt}

\begin{prompt}[{Python Code for ImagePatch Class}]
\begin{lstlisting}[language=Python]
class ImagePatch:
    def __init__(self, image, left=None, lower=None, right=None, upper=None):
        self.image 
        pass
        
    @property
    def area(self):
        pass

    def find(self, object_name):
        pass

    def exists(self, object_name):
        pass

    def verify_property(self, object_name, visual_property):
        pass

    def best_description_from_options(self, object_name, property_list):
        pass

    def simple_query(self, question):
        pass

    def crop_left_of_bbox(self, left, upper, right, lower):
        pass

    def crop_right_of_bbox(self, left, upper, right, lower):
        pass

    def crop_below_bbox(self, left, upper, right, lower):
        pass

    def crop_above_bbox(self, left, upper, right, lower):
        pass

    def llm_query(self, question):
        pass
        
def bool_to_yesno(bool_answer: bool) -> str:
    pass
    
\end{lstlisting}
\end{prompt}
